\pdfoutput=1
\documentclass[11pt]{article}

\usepackage{xcolor}
\usepackage{listings}
\usepackage{courier}

\lstdefinelanguage{json}{
    basicstyle=\ttfamily,
    numbers=left,
    numberstyle=\tiny,
    stepnumber=1,
    numbersep=8pt,
    showstringspaces=false,
    breaklines=true,
    frame=single,
    backgroundcolor=\color{gray!10},
    stringstyle=\color{blue},
    keywordstyle=\color{red},
    morestring=[b]",
    morekeywords={true,false,null}
}

\usepackage[preprint]{acl}

\usepackage{booktabs}

\usepackage{amsmath}

\usepackage{makecell}

\usepackage{times}
\usepackage{latexsym}
\usepackage{multirow}

\usepackage[T1]{fontenc}
\usepackage[utf8]{inputenc}
\usepackage[utf8]{inputenc}

\usepackage{microtype}

\usepackage{inconsolata}

\usepackage{graphicx}

\usepackage{comment}
\usepackage[colorinlistoftodos]{todonotes}

\usepackage[normalem]{ulem}

\usepackage{fancyhdr}

\title{Cheaper, Better, Faster, Stronger: \\ Robust Text-to-SQL without Chain-of-Thought or Fine-Tuning}

\author{
\textbf{Yusuf Denizay Dönder}$^{1}$\thanks{Equal contribution.} ,
\textbf{Derek Hommel}$^{1}$\footnotemark[1] ,
\textbf{Andrea W Wen-Yi}$^{2}$, \\
\textbf{David Mimno}$^{2}$, \textbf{Unso Eun Seo Jo}$^{1, 2}$ \\
$^{1}$Gena Co.; $^{2}$Cornell University\\
\texttt{unsojo@cornell.edu}
}

\begin{document}
\maketitle

\pagestyle{fancy}
\fancyhf{}  % Clear existing headers/footers
\fancyhead[L]{Preprint. Under review.}  % Centered header

\begin{abstract}
LLMs are effective at code generation tasks like text-to-SQL, but is it worth the cost? Many SOTA approaches use non-task-specific LLM approaches including Chain-of-Thought (CoT), self-consistency, and fine-tuning models. These methods can be costly for inference, sometimes requiring over a hundred LLM calls with reasoning, incurring average costs of up to \$0.46 per query, while fine-tuning models can cost up to thousands of dollars. We introduce \textit{“N-rep”} consistency, a more cost efficient text-to-SQL approach that achieves similar BIRD benchmark scores as other more expensive methods, only costing \$0.039 per query. \textit{N-rep} leverages multiple representations of the same schema input to mitigate weaknesses in any single representation, making the solution more robust and allowing the use of smaller and cheaper models without any reasoning or fine-tuning. To our knowledge, \textit{N-rep} is the best performing text-to-SQL approach in its cost range.
\end{abstract}

\section{Introduction}

\begin{comment}

\end{comment}
While LLMs have boosted SOTA performance on enduring code-generation tasks such as text-to-SQL, these approaches remain well below human-level performance \cite{li2024can}.

Current LLM text-to-SQL approaches rely on two computationally expensive techniques: Chain-of-Thought (CoT) self-consistency \cite{52081} and fine-tuning \cite{gao2025previewxiyansqlmultigeneratorensemble}. Chain-of-Thought self-consistency uses CoT prompting to generate a pool of candidate queries, requiring many LLM calls \cite{talaei2024chesscontextualharnessingefficient}. Fine-tuning requires extensive resources for data labeling and model training. Combining both approaches amplifies the costs.

\begin{figure}[t]
    \centering
    \includegraphics[width=0.48\textwidth]{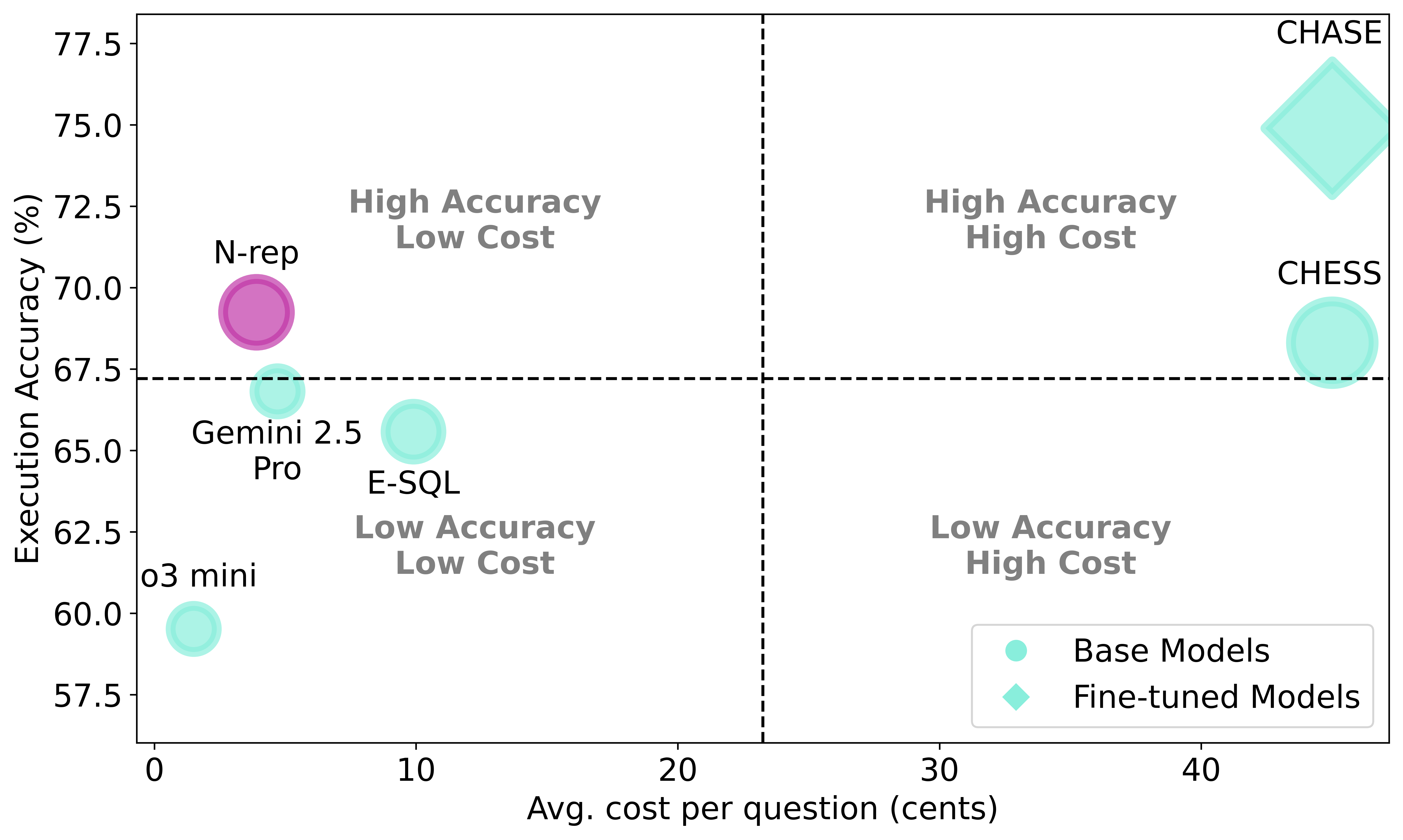}
    \caption{Comparison of Execution Accuracy (EX) and average cost (cents) per query for different models on the BIRD benchmark dev set. Shape sizes reflect the logarithmic scale of the number of LLM calls.}
    \label{fig:efficiency-scatter}
\end{figure}

\begin{figure*}[t]
    \centering
    \includegraphics[width=\textwidth]{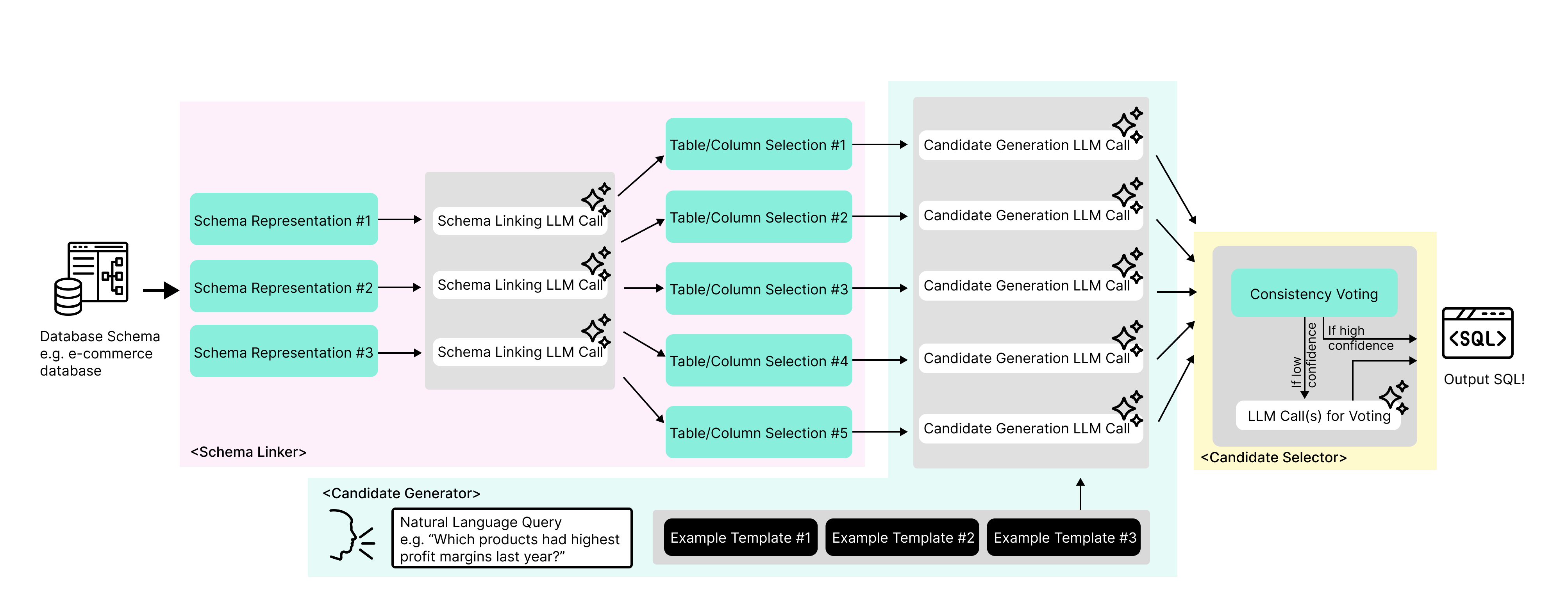}
    \caption{Overview of the N-rep approach for Text-to-SQL generation.}
    \label{fig:mvc-overview}
\end{figure*}
\begin{table*}[t]
    \centering
    \begin{tabular}{lcccc}
    \toprule
    % \textbf{Model} & \textbf{Execution Accuracy} & \textbf{LLM Calls Typical (Avg.)} & \textbf{Token Count (K)} & \textbf{Cost (\$)} \\
    \textbf{Model} & \textbf{Execution Accuracy} & \textbf{LLM Calls Typical(Avg.)} & \textbf{Tokens (K)} & \textbf{Cost (\$)} \\
    \midrule
    \multicolumn{5}{l}{\textbf{Qwen3\textsuperscript{‡}}} \\
    \quad 8B | 14B | 32B   & 44.98 | 53.46 | 56.71 & 1(1) | 1(1) | 1(1) & 3.2 | 3.2 | 3.2 & --- \\
    \midrule
    \multicolumn{5}{l}{\textbf{N-rep w/ Qwen3\textsuperscript{‡}}} \\
        \quad 8B | 14B | 32B   & 57.04 | 61.67 | 64.02 & 8(12.6) | 8(12.0) | 8(10.6) & 36.8 | 33.9 | 31.7 & --- \\
    \midrule
    o3-mini       & 59.52 & 1(1) & 3.8 & 0.015 \\
    \textbf{N-rep (ours)} & \textbf{69.25} & \textbf{8(10.9)} & \textbf{32.0} & \textbf{0.039} \\
    Gemini 2.5 Pro & 66.82 & 1(1) & 4.7 & 0.047 \\
    E-SQL         & 65.58 & 3(3) & 3.7 & 0.09 \\
    CHASE\textsuperscript{*} & 74.9 & --- (100+) & --- & 0.44 \\
    CHESS         & 68.31 & --- (31.83) & 319.9 & 0.46 \\
    \bottomrule
    \end{tabular}
    \caption{
    Execution Accuracy (EX) comparison of models along with average per query LLM
    calls, token usage  (in thousands), and cost on the BIRD dev set. \textit{Typical} shows median and the \textit{Avg.} shows the mean number of calls. Same prompt with 3 few-shot samples used for base models without frameworks. Refer to Appendix~\ref{sec:method-for-analysis} for details on configurations and cost calculations.
    \textsuperscript{‡}Open-source model. 
    \textsuperscript{*}Model uses fine-tuning.
    }
    \label{tab:model-efficiency}
\end{table*}

We introduce ``N-rep'' Consistency, a novel, cheaper, SQL-specific approach leveraging $n$ different text representations of the database schema for both schema linking and candidate generation. By utilizing various schema representations with different schema linking filtering, our method is robust to input sensitivity and schema linking errors while using fewer candidates than other consistency-based approaches. ``N-rep'' requires fewer tokens per candidate than reasoning methods and performs comparably to top-10 ranking approaches on the BIRD benchmark while costing nearly 10 times less per query, without fine-tuning expenses.\footnote{\url{bird-bench.github.io} as of May 2025} Open source package will be available upon acceptance.

\section{Related Work}

\paragraph{Schema Linking}
\textit{Schema Linking} is the task of identifying the database tables and columns relevant to the user’s question. \citet{maamari2024death} leverage LLM long context windows to bypass a schema linking step. \citet{qu-etal-2024-generation} utilize SQL generation for schema linking by prompting an LLM to generate an SQL query using the full schema, then using the tables and columns from that prediction as inputs for the final SQL generation. Recent approaches utilize more elaborate multi-step schema linking methods that use locality-sensitive hashing (LSH) to check keywords against database values and leverage external column description information when available \cite{talaei2024chesscontextualharnessingefficient, pourreza2025chasesql, gao2025previewxiyansqlmultigeneratorensemble}.

\paragraph{Schema Representation}
How to format database schemas for LLMs is an area of active research. DAIL-SQL \cite{gao2023texttosqlempoweredlargelanguage}, DTS-SQL \cite{pourreza-rafiei-2024-dts}, CHESS \cite{talaei2024chesscontextualharnessingefficient} and E-SQL \cite{caferouglu2024sql} use data description language (DDL); TA-SQL \cite{qu-etal-2024-generation} use a dictionary with \texttt{table.column} keys and column description values; \newcite{maamari2024death} use bulleted lists of tables, columns, types and sample values; DIN-SQL \cite{10.5555/3666122.3667699} lists each table on one line with a list of its columns. MAC-schema \cite{wang-etal-2025-mac} and M-Schema \cite{gao2025previewxiyansqlmultigeneratorensemble} are semi-structured representations, with each table shown as a list of column tuples. Both MAC-schema and M-schema outperform the DDL format on various LLMs \cite{gao2025previewxiyansqlmultigeneratorensemble}. See Appendix~\ref{sec:schema-format} for examples.

\paragraph{Self-consistency}
CoT self-consistency \cite{52081} combines CoT prompting \cite{10.5555/3600270.3602070} and temperature sampling to generate candidate answers from multiple reasoning paths before selecting the most ``consistent'' one. CHASE-SQL \cite{pourreza2025chasesql} utilizes \textit{multi-path candidate generation} with multiple CoT templates to generate 21 candidates, and uses a fine-tuned LLM to select the final SQL via pairwise comparison. CHESS generates 20 candidates via a structured CoT prompt and uses voting that evaluates each candidate against LLM-generated ``unit tests''. XiYan-SQL \cite{gao2025previewxiyansqlmultigeneratorensemble} utilizes multiple fine-tuned LLMs to generate 5 candidates, then another fine-tuned LLM for candidate selection. 

\section{Motivation} 
Our approach is motivated by a key observation: LLMs are sensitive to schema representation formats \cite{gao2025previewxiyansqlmultigeneratorensemble}. Varying schema representations creates meaningful candidate diversity without temperature sampling or CoT reasoning. Generating multiple ``schema linking candidates'' is also simpler and more robust than other approaches that rely on extensive preprocessing and multiple LLM calls that are prone to error propagation from missing critical tables/columns. 

\section{Methodology}

\noindent\textbf{The N-rep Framework} has three stages (Figure~\ref{fig:mvc-overview}). \textbf{Schema Linker} utilizes multiple schema representations for schema linking. The results get passed to \textbf{Candidate Generator} to create multiple candidate SQL queries. Finally, the \textbf{Candidate Selector} uses a two-stage confidence-aware selection process to choose the final SQL query.

\subsection{Schema Linker}
\label{subsec:schema-linker}

This stage identifies tables and columns required to answer the natural language question (NLQ). First, target database schema is converted into $n$ representations. Second, each representation, along with the NLQ and a fixed set of 3 hand-selected few-shot examples, is used to predict relevant tables and columns. 
Lastly, we produce three schema linking results for each representation --- one using the full database schema (\textit{no filtering}), one with only the selected tables and all their columns (\textit{table-only}), and one with only the selected tables and columns (\textit{full filtering}). See Appendix~\ref{sec:filter-level} for illustrations.

\subsection{Candidate Generator}
\label{subsec:candidate-generator}
This stage generates multiple SQL candidates. An LLM takes a schema representation from previous stage --- either fully filtered, table-only, or unfiltered ---the NLQ, and 3 few-shot examples to generate a candidate SQL. N-rep retrieves few-shot samples with fully-filtered schema information from the training dataset following XiYan-SQL.  See Appendix~\ref{subsec:generator-system-prompt} for the system prompt.

\subsection{Candidate Selector}
N-rep adopts a \textbf{confidence-aware} two-stage candidate selection strategy that combines regular self-consistency voting with CHASE-SQL's LLM-based pair-wise voting.\footnote{In some cases, the LLM based selection is more effective, but it may require up to $2 \cdot \binom{M}{2}$ LLM calls. Each pairwise comparison is done twice by swapping the order to mitigate any order bias.} Using the number of votes from regular voting as an indicator of confidence, N-rep applies LLM-based selection only for low confidence cases. Appendix~\ref{subsec:candidate-select-detail} shows the details for confidence threshold and the Appendix~\ref{subsec:ex-by-vote} shows the relation between confidence and performance.

\label{subsec:candidate-selection}

\section{Experiments}

\begin{figure}[t]
    \centering
    \includegraphics[width=0.48\textwidth]{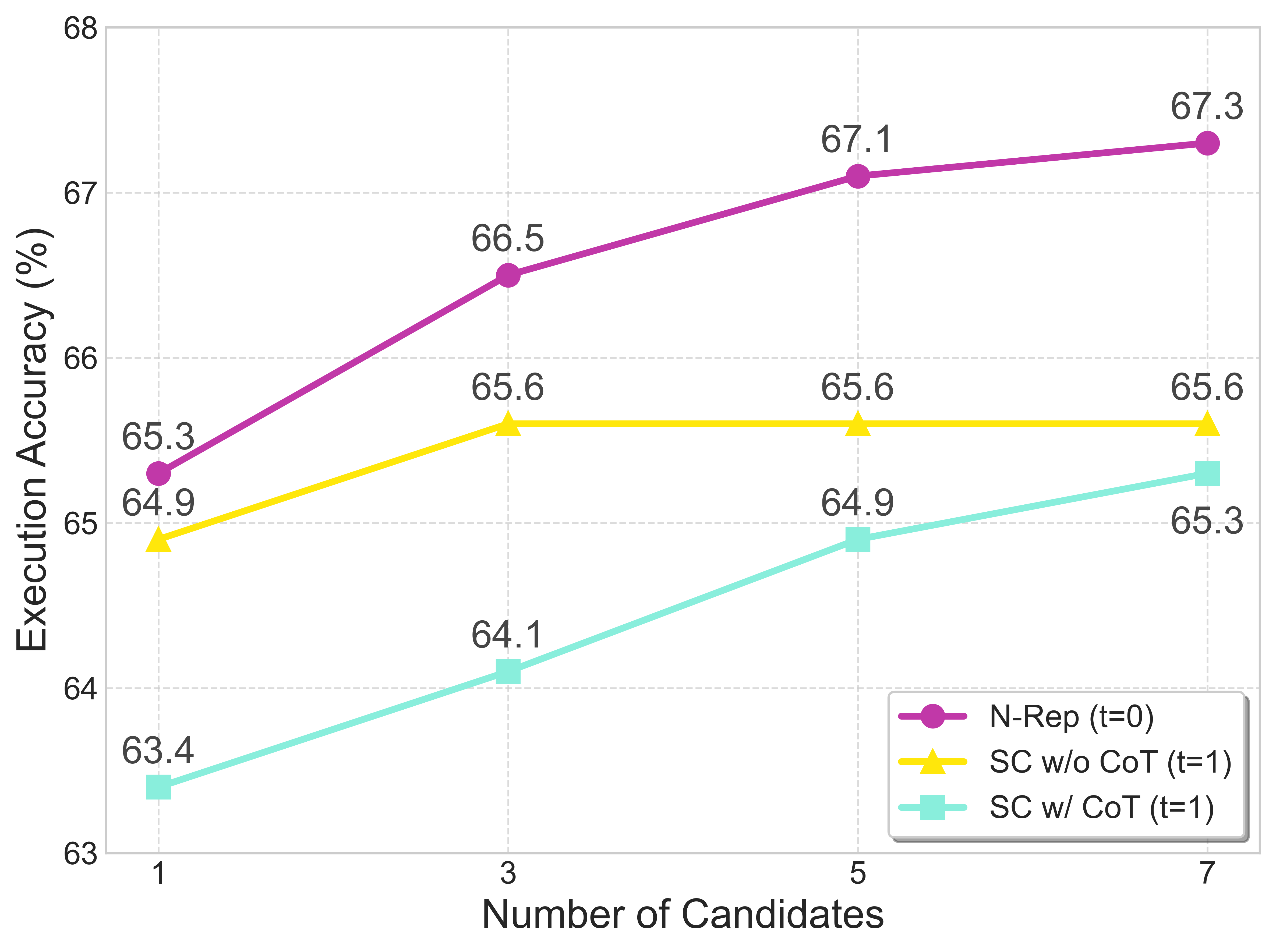}
     \caption{Comparison of N-rep, Self Consistency with CoT and Self Consistency without CoT. $t=1$ means sampling temperature is 1.}
    \label{fig:sel-acc-comparison}
\end{figure}

\subsection{Datasets and Metrics}
We evaluate the N-rep framework on the BIRD \cite{li2024can} and SPIDER \cite{yu-etal-2018-spider} benchmarks. SPIDER is a cross-domain benchmark with 200 databases and 10,181 query pairs (7000 train, 1034 dev, 2147 test). BIRD contains 95 databases and 12,751 query pairs (9428 train, 1534 dev, 1789 test) that covers more than 37 real-world domains such as finance and medicine. For evaluation, we use Execution Accuracy (EX), the average number of queries where the predicted and ground truth SQL execution results are identical. 

We compare N-rep (Table~\ref{tab:model-efficiency}) to other publicly available solutions, and a selection of closed and open-source base models, including \texttt{OpenAI o3-mini} \cite{o3-mini}, \texttt{Gemini 2.5 pro} \cite{gemini-25}, and \texttt{Qwen3} \cite{yang2025qwen3technicalreport}. Base model tests use 3 few-shot samples from train sets following XiYan-SQL.

\subsection{Experiment Setup}
To achieve our best performing configuration of N-rep, we select 4 best schema representations based on schema linking experiments to optimize for precision-recall tradeoffs. For each candidate size $n$ in Figure~\ref{fig:sel-acc-comparison}, we conduct a parameter search across all combinations of our 4 selected representations and 3 table filtering levels on a 10\% dev subset, using regular voting to identify the highest-EX combination. All experiments in \ref{sc-vs-n-rep} and \ref{subsec:candidate-selection-experiments} use \texttt{Gemini 1.5 Flash} \cite{team2024gemini} with 3 few-shot samples. Baseline and test configurations are described in Appendix~\ref{sec:method-for-analysis}, and final N-rep configuration is shown in Appendix~\ref{subsec:final-config}. 

\subsection{Self-Consistency vs N-rep}
\label{sc-vs-n-rep}
We compare the performance of N-rep with greedy decoding against traditional self-consistency using temperature sampling with and without CoT in Figure~\ref{fig:sel-acc-comparison}.
While EX of self-consistency without CoT plateaus after 3 candidates, EX of CoT self-consistency and N-rep both increase with the number of candidates. Moreover, for all candidate numbers tested, EX of N-rep is 2 percentage points higher than that of CoT self-consistency. Furthermore, Figure~\ref{fig:acc-bounds} shows that N-rep's upper bound increases more than CoT self-consistency's as candidates increase, suggesting that with better selection methods, N-rep's performance over CoT self-consistency could further increase.

\subsection{Candidate Selector}
\label{subsec:candidate-selection-experiments}
We compare the regular voting system, LLM-based voting system and our \textit{confidence-aware} method. Table~\ref{tab:select-method} shows confidence-aware final SQL selection can reduce the number of LLM calls by 60\% while increasing the EX. 

\begin{table}[h]
\centering
\begin{tabular}{lcc }
\toprule
Selection Method & EX & LLM Calls \\
\midrule
Regular Voting    	& 67.2 & 0\\
LLM Based Voting	& 67.9 & 6178\\
Confidence-aware Voting     		& 68.8 & 2436\\
\bottomrule
\end{tabular}
\caption{Execution Accuracy (EX) and number of LLM calls over the whole BIRD development set for different selection methods. }
\label{tab:select-method}
\end{table}

\begin{table}[t]
\centering
\begin{tabular}{lc}
\toprule
Model & Execution Accuracy \\
\midrule
MiniSeek	& 91.2\\
CHESS       & 87.2 \\
N-Rep (Ours)       & 87.0 \\
DAIL-SQL    & 86.6 \\
\bottomrule
\end{tabular}
\caption{Execution Accuracy (EX) comparison of models on the SPIDER test set.}
\label{tab:spider-results}
\end{table}

\subsection{Cost Analysis}

Compared to \texttt{Gemini 2.5 Pro} baseline, N-rep achieves 2.43 percentage points higher EX at a 17\% cheaper price per query (\$0.039 vs \$0.047),  Against CHESS, a CoT self-consistency method, N-Rep scores 0.94 higher EX while using only a third of LLM calls and 8.5\% of the inference cost; against CHASE, a leading CoT self-consistency method using a fine-tuned candidate selection LLM, N-rep comes within 5.65 percentage points EX at only 8.7\% of the inference cost (\$0.039 vs \$0.44) per query (Table~\ref{tab:model-efficiency}).

\section{Discussion}

Our findings reveal that LLMs remain sensitive to schema representation for text-to-SQL, suggesting the models rely more on surface understanding than deep reasoning about database structures. The N-rep approach leverages this insight by combining multiple representations.

Our experiments with the \texttt{Qwen3} models suggest that \textit{N-rep reduces reliance on model scale}. Specifically, smaller models (e.g., \texttt{Qwen3-8B}) exhibit larger relative gains in EX when augmented with N-rep --- improving from 44.98\% to 57.04\% EX, a jump of over 12 points --- compared to a larger model like \texttt{Qwen3-32B}, which improved by 7.3 percentage points (Table~\ref{tab:model-efficiency}).

To evaluate generalizability, we also tested N-rep on the SPIDER benchmark (Table~\ref{tab:spider-results}), where it achieves 87.0 execution accuracy indicating that our solution is robust across benchmarks.

These results show how domain expertise — specifically, understanding the impact of schema representation variability — can outperform generic approaches like CoT self consistency that do not address this limitation (Figure~\ref{fig:sel-acc-comparison}).

\section{Conclusion}
We propose N-rep, a robust and cheaper framework that utilizes multiple text representations of database schemas. N-rep serves as a case study of how, for specific well-defined tasks with fixed input and output, using domain specific knowledge can minimize cost of LLM-based solutions. 

\section*{Limitations}
The requirement for using manually-selected schema representations limits the number of total candidates with unique representations, and our analysis is not exhaustive. Future work could examine a broader range of schema representations, and explore generating alternate forms with LLMs.

Also, our confidence-aware voting policy is currently hand-tailored for each specific number of candidates. Future work could optimize and automate this voting policy to scale to an arbitrary number of candidates.

\section*{Acknowledgments}
\textit{Left blank for anonymous submission.}
\bibliography{anthology,custom}

\appendix

\section{Methodology for performance and cost analysis}
\label{sec:method-for-analysis}
\subsection{Models Chosen}
For our comparison, along with our main N-rep solution, we included results for: three frameworks from the BIRD leaderboard, two closed-source models, and three versions of the \texttt{Qwen3} family of models with and without our N-rep framework.

We selected CHESS and E-SQL because their authors shared code, enabling us to replicate results and measure token usage. For these, we used our own LLM call and token measurements, reporting EX values as published.

For CHASE, although the code was not available we referenced a graph that includes cost information in another paper \newcite{pourreza2025reasoning} to estimate the cost.

We included \texttt{OpenAI o3-mini} and \texttt{Gemini 2.5 Pro} to evaluate out-of-box capabilities of closed-source reasoning models. The \texttt{Qwen3} series was selected specifically because multiple model sizes are available, allowing us to analyze scaling in performance with our method.

\subsection{Configurations}
For \texttt{Gemini 2.5 Pro} and \texttt{o3-mini}, we conducted experiments using default settings with reasoning enabled. For the \texttt{Qwen3} models, we disabled reasoning capabilities to ensure fair comparison between the base implementation and our N-rep approach.
Both the closed-source models and the \texttt{Qwen3} models without N-rep used identical prompts with three few-shot samples for consistency, and the temperature was set to 0.

We use Cohere Multilingual v3 with "clustering" input type \cite{cohere-embed} for all embeddings for few-shot retrieval. 

All experimental results reported are based on a single run.

\subsection{Cost Calculation}
We calculated costs using current pricing models as of May 16, 2025. For the o3-mini experiments, we used Azure OpenAI Service pricing of \$1.10 per 1M input tokens and \$4.40 per 1M output tokens. For E-SQL, which used \texttt{GPT-4o} \cite{hurst2024gpt}, \$2.50 per 1M input tokens and \$10.00 per 1M output tokens. 

For \texttt{Gemini 2.5 Pro} and CHESS, which uses \texttt{Gemini 1.5 Pro}, we applied Gemini Developer API pricing: \$1.25 per 1M input tokens and \$10.00 per 1M output tokens.

All calculations reflect full pricing without accounting for potential discounts from cache hits, providing a consistent basis for comparison across all models.

\section{Implementation details}
\subsection{Final configuration}
\label{subsec:final-config}

\begin{table}[h]
\centering
\begin{tabular}{ccc}
\toprule
\textbf{Format} & \textbf{Filtering level} & \textbf{Model} \\
\midrule
MAC Schema & No Filtering & — \\
MAC Schema & Col. Filtering & GPT-4o \\
M-Schema & Table Filtering & GPT-4o \\
M-Schema & Full Filtering & GPT-4o \\
DDL & Full Filtering & \makecell{Gemini 1.5\\Pro} \\
\bottomrule
\end{tabular}
\caption{Schema representations used in the final implementation. Note that the rows for \textit{M-Schema with Table Filtering} and \textit{M-Schema with Full Filtering} use the same prediction, corresponding to a single LLM call.}
\label{tab:schema_representations}
\end{table}

\begin{table*}[t]
\centering
\begin{tabular}{lcccccc}
\hline
\multirow{2}{*}{Format} & \multicolumn{3}{c}{Table} & \multicolumn{3}{c}{Column} \\
\cline{2-7}
 & Precision & Recall  & F1 & Precision & Recall & F1 \\\hline
MAC-Schema         & 92.9 & 93.1 & 93.0 & 78.1 & 87.2 & 82.4 \\
SQL Alchemy        & 91.7 & 93.2 & 92.5 & 70.7 & 87.2 & 78.1 \\
JSON Raw           & 91.8 & 92.9 & 92.4 & 74.9 & 87.2 & 80.6 \\
DIN-SQL            & 90.8 & 93.2 & 92.0 & 74.9 & 86.9 & 80.4 \\
M-Schema           & 91.8 & 93.1 & 92.5 & 75.5 & 87.1 & 80.9 \\
DDL                & 92.2 & 93.6 & 92.8 & 75.5 & 86.9 & 82.5   \\
\hline
\end{tabular}
\caption{Micro precision and recall scores (in \%) for different formats with GPT-4o}
\label{tab:schema-linking}
\end{table*}

\begin{table*}[t]
\centering
\begin{tabular}{lcccccc}
\hline
\multirow{2}{*}{Format} & \multicolumn{3}{c}{Table} & \multicolumn{3}{c}{Column} \\
\cline{2-7}
 & Precision & Recall  & F1 & Precision & Recall & F1 \\\hline
MAC-Schema         & 71.5 & 96.1 & 82.0 & 37.6 & 94.0 & 53.7  \\
M-Schema           & 68.1 & 95.4 & 79.5 & 35.6 & 92.4 & 51.4  \\
DDL                & 69.2 & 95.7 & 80.3 & 39.2 & 92.7 & 55.1  \\
\hline
\end{tabular}
\caption{Micro precision and recall scores (in \%) for different formats with Gemini 1.5 Pro}
\label{tab:schema-linking-gemini}
\end{table*}

For our final configuration of N-rep, we used 5 candidates. \texttt{Gemini 1.5 Flash} was used for all candidate generations and LLM-based candidate selections for low confidence answers. 

For the schema linking phase we used three formats: DDL format, MAC-Schema format, and M-Schema format, making 3 calls in total for the whole phase. We also used 2 different models to further increase the variety of representations: \texttt{Gemini 1.5 Pro} and \texttt{GPT-4o}. We used different levels of filtering, again, to increase the range of representations. Table~\ref{tab:schema_representations} shows the specific representations used for our configuration. The temperature is set to 0 for all models.

\subsection{Candidate selection}
\label{subsec:candidate-select-detail}
In cases where vote distribution indicates low confidence we apply LLM-based comparison.
In our five-candidate setup, we invoke the LLM selector when: all candidates receive only one vote, two candidates each receive two votes (indicating a tie among plausible options), the top has three votes, but another has two (indicating a strong second contender).

\section{Detailed Results}
\label{sec:additional-experiments}
\subsection{Schema Linking}
\label{subsec:diff-format-comparisons}

Table~\ref{tab:schema_representations} shows table and column performance on schema linking task for our six candidate schema representations. After selecting the best three formats by column F1, we did further tests with \texttt{Gemini 1.5 Pro} (Table~\ref{tab:schema-linking-gemini}).

\subsection{Execution Accuracy Upper \& Lower Bounds}

\begin{figure}[h]
    \centering
    \includegraphics[width=0.48\textwidth]{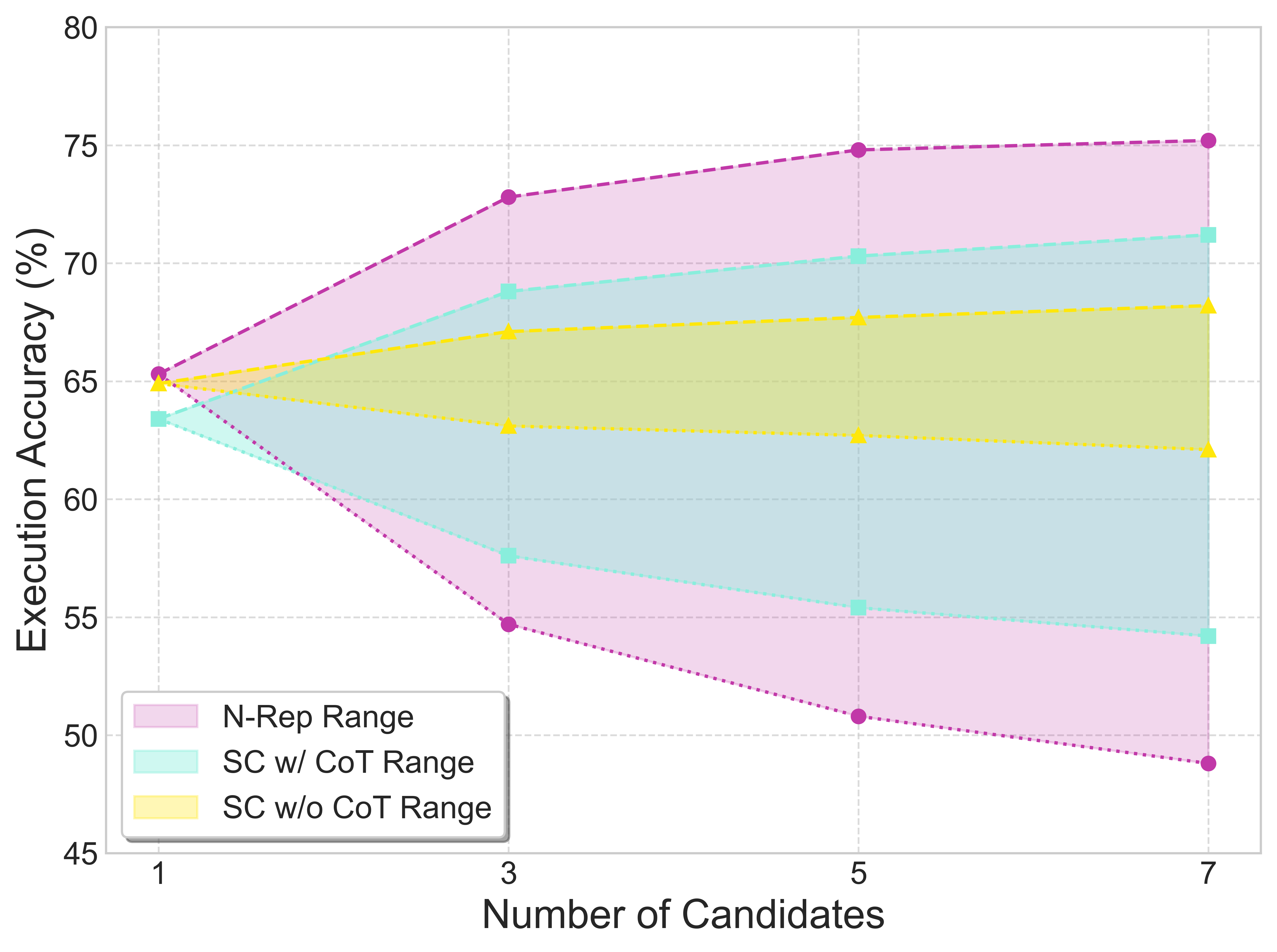}
     \caption{Upper and lower bounds of N-rep, Self Consistency with CoT and Self Consistency without CoT.}
    \label{fig:acc-bounds}
\end{figure}

Figure~\ref{fig:acc-bounds} shows upper and lower bounds for N-rep vs self-consistency for 1, 3, 5 and 7 candidates. The \textit{upper bound} reflects the best-case scenario where a perfect candidate selector chooses the correct SQL query from the set, i.e. at least one candidate is correct. The \textit{lower bound} represents the worst-case selection, i.e. at least one candidate is wrong.
The chance of at least one being wrong goes up as the number of candidates goes up and the chance of at least once being right goes up as the number of candidates goes up.

\label{subsec:acc-bounds}

\label{subsec:ex-by-vote}
\begin{figure*}[h]
    \centering
    \includegraphics[width=\textwidth]{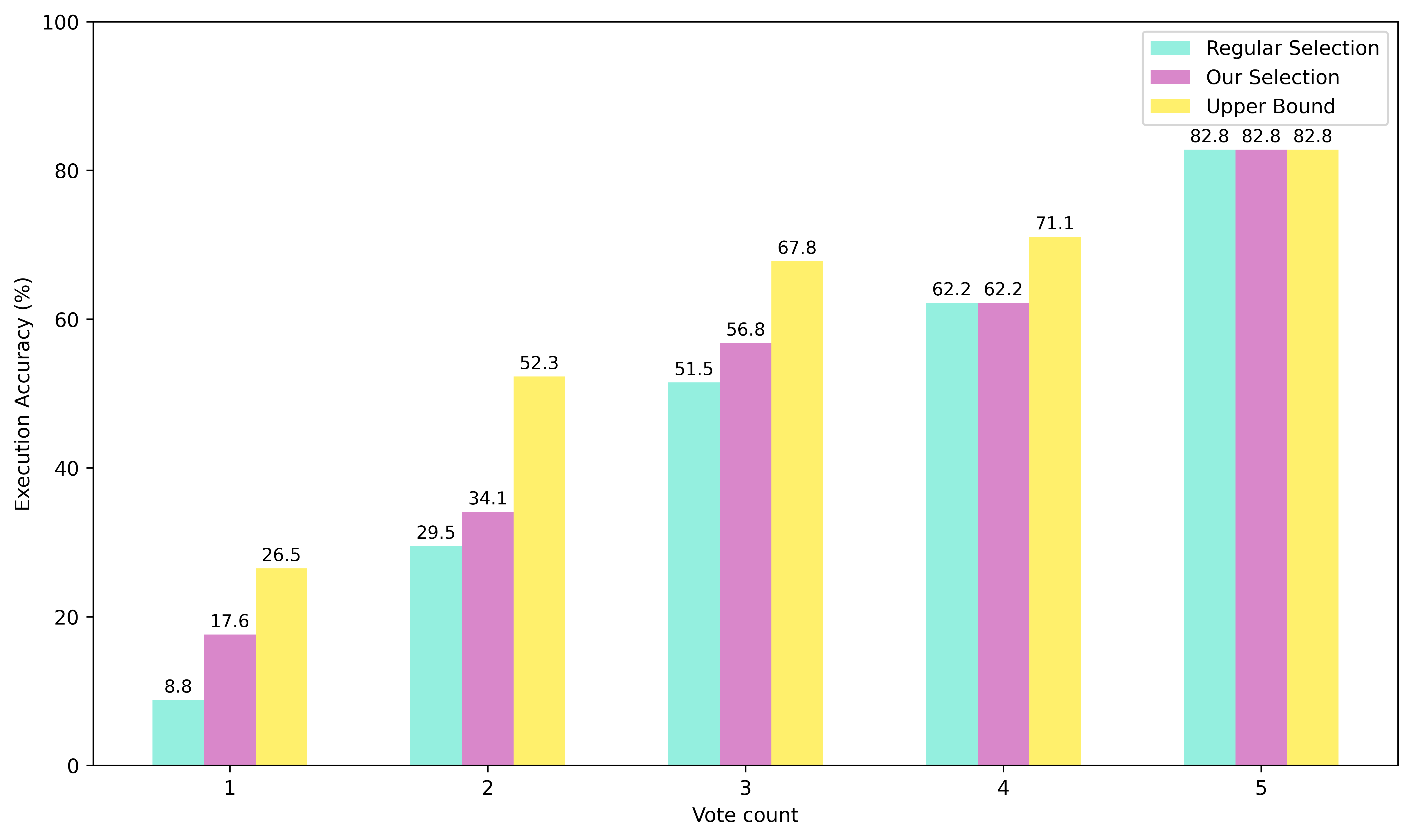}
    \caption{EX by vote count for the selected candidate}
    \label{fig:ex-by-vote}
\end{figure*}

\subsection{Execution Accuracy by Vote Count}

As seen in Figure~\ref{fig:ex-by-vote}, the upper bound values increase with vote count, demonstrating that the number of votes assigned to the winning group corresponds to the likelihood that at least one candidate is correct, thus, higher confidence. The upper bound shows the maximum EX assuming perfect candidate selection. Regular selection and our confidence-aware selection are equivalent at count = 4; all three are all equal in the case of count = 5 because all candidates have equivalent output.

\onecolumn

\section{Prompt Details}
\label{sec:prompt-details}

\subsection{Candidate Generator System Prompt}
\label{subsec:generator-system-prompt}

Figure~\ref{fig:system-prompt} shows the system prompt for candidate generation.

\begin{figure*}[h!]
  \centering
  \input{candidate_generator_system_prompt}
  \caption{candidate generator system prompt}
  \label{fig:system-prompt}
\end{figure*}

\clearpage
\section {Schema Formats}
\label{sec:schema-format}

We test six schema representations, M-Schema (Figure~\ref{fig:format-mschema}, MAC-Schema (Figure~\ref{fig:format-mac}, DDL (Figure~\ref{fig:format-ddl}, DIN-SQL (Figure~\ref{fig:format-dinsql}, a raw JSON representation (Figure~\ref{fig:format-json}, and a python representation (Figure~\ref{fig:format-sqlalchemy}.

\begin{figure*}[h!]
  \centering
  \input{sample_m_schema}
  \caption{XiYan-SQL M-SCHEMA format}
  \label{fig:format-mschema}
\end{figure*}

\begin{figure*}[h!]
  \centering
  \input{sample_mac_schema}
  \caption{MAC-SCHEMA format}
  \label{fig:format-mac}
\end{figure*}

\begin{figure*}[h!]
  \centering
  \input{sample_sql_create}
  \caption{DDL ("SQL CREATE") format}
  \label{fig:format-ddl}
\end{figure*}

\begin{figure*}[h!]
  \centering
  \input{sample_din}
  \caption{DIN-SQL style format}
  \label{fig:format-dinsql}
\end{figure*}

\begin{figure*}[h!]
  \centering
  \input{sample_json_raw}
  \caption{raw JSON format}
  \label{fig:format-json}
\end{figure*}

\begin{figure*}[h!]
  \centering
  \input{sample_sql_alchemy}
  \caption{python SQLAlchemy format}
  \label{fig:format-sqlalchemy}
\end{figure*}

\clearpage
\section {Filtering Levels}

This shows examples of different schema filtering levels on a toy example dataset in the compact "DIN-SQL" format. The toy dataset has three tables in total, "users", "orders" and "products" (Figure~\ref{fig:no-linking-example}). The example schema linking prediction output is a JSON-like dictionary; here it predicted we need to keep two tables, "users" and "orders", with two columns each (Figure~\ref{fig:linking-inputs}). \textit{No filtering} preserves all schema entities (Figure~\ref{fig:no-linking-example}); \textit{table-only filtering} drops the "products" table, but all columns from "users" and "orders" remain (Figure~\ref{fig:table-linking-example}); \textit{full filtering} removes all entities that are not in the prediction (Figure~\ref{fig:full-linking-example}).

\begin{figure*}[h!]
  \centering
  \input{filtering_input_sample}
  \caption{Schema linker output}
  \label{fig:linking-inputs}
\end{figure*}

\label{sec:filter-level}

\begin{figure*}[h!]
  \centering
  \input{no_linking_sample}
  \caption{No filtering example (full schema)}
  \label{fig:no-linking-example}
\end{figure*}

\begin{figure*}[h!]
  \centering
  \input{table_linking_sample}
  \caption{Table-only filtering example}
  \label{fig:table-linking-example}
\end{figure*}

\begin{figure*}[h!]
  \centering
  \input{full_linking_sample}
  \caption{Full filtering example}
  \label{fig:full-linking-example}
\end{figure*}

\twocolumn

\end{document}